\documentclass[letterpaper, 10 pt, conference]{ieeeconf}  

\IEEEoverridecommandlockouts                              

\makeatletter
\def\endthebibliography{%
	\def\@noitemerr{\@latex@warning{Empty `thebibliography' environment}}%
	\endlist
}
\makeatother

\usepackage{cite}
\usepackage{amsmath,amssymb,amsfonts}

\usepackage{graphicx}
\usepackage{textcomp}
\usepackage{xcolor,soul}

\usepackage{algpseudocode}
\usepackage{algorithm}

\usepackage{mathtools}

\usepackage{mathrsfs}

\usepackage{todonotes}
\usepackage{xargs}
\usepackage{svg}
\usepackage{booktabs}
\usepackage{multirow}

\newtheorem{remark}{\textbf{Remark}}

\newtheorem{assumption}{\textbf{Assumption}}
\newtheorem{definition}{\textbf{Definition}}

\newtheorem{example}{\textbf{Example}}

\renewcommand{\baselinestretch}{0.895}

\title{\LARGE \bf
	A Counterfactual Reasoning Framework for Fault Diagnosis\\ in Robot Perception Systems}

\author{Haeyoon Han$^{*1}$\thanks{$^{*}$Haeyoon Han and Mahdi Taheri are co-first authors.}, Mahdi Taheri$^{*1}$, Soon-Jo Chung$^{1}$, and Fred Y. Hadaegh$^{1}$
    \thanks{$^{1}$Division of Engineering and Applied Science, California Institute of Technology (Caltech), Pasadena, CA 91125, \{hhan3, mtaheri, sjchung, hadaegh\}@caltech.edu.}%
}

\begin{document}

	\maketitle
	\thispagestyle{empty}
	\pagestyle{empty}
	
	\begin{abstract}
    Perception systems provide a rich understanding of the environment for autonomous systems, shaping decisions in all downstream modules. Hence, accurate detection and isolation of faults in perception systems is important. Faults in perception systems pose particular challenges: faults are often tied to the perceptual context of the environment, and errors in their multi-stage pipelines can propagate across modules. To address this, we adopt a counterfactual reasoning approach to propose a framework for fault detection and isolation (FDI) in perception systems. As opposed to relying on physical redundancy (i.e., having extra sensors), our approach utilizes analytical redundancy with counterfactual reasoning to construct perception reliability tests as causal outcomes influenced by system states and fault scenarios. Counterfactual reasoning generates reliability test results under hypothesized faults to update the belief over fault hypotheses. We derive both passive and active FDI methods. While the passive FDI can be achieved by belief updates, the active FDI approach is defined as a causal bandit problem, where we utilize Monte Carlo Tree Search (MCTS) with upper confidence bound (UCB) to find control inputs that maximize a detection and isolation metric, designated as Effective Information (EI). The mentioned metric quantifies the informativeness of control inputs for FDI. We demonstrate the approach in a robot exploration scenario, where a space robot performing vision-based navigation actively adjusts its attitude to increase EI and correctly isolate faults caused by sensor damage, dynamic scenes, and perceptual degradation. 
	\end{abstract}
	\section{Introduction}
    
    Autonomous systems such as self-driving cars, unmanned aerial vehicles (UAV), and autonomous robots rely on perception systems to convert heterogeneous sensor measurements into a coherent representation of their surrounding environment~\cite{corke_robotics_2017}. The role of the perception system is to provide accurate and timely information on objects, terrain, and the surrounding environment so that higher-level modules in an autonomous system (e.g., localization, motion planning, and control) can guarantee safety and achieve mission objectives~\cite{sinha_closing_2023}. The combination of utilizing heterogeneous sensors (e.g., LiDAR, radar, cameras) and deep learning-based algorithms has led to recent advances in perception-based control. However, this has also resulted in an increased level of complexity in perception systems, which makes detecting their faults and algorithmic errors challenging~\cite{antonante_monitoring_2023,abdul_hafez_quantifying_2024}. Considering the importance of a perception system in the guidance and control of an autonomous system, perception faults can result in the complete loss of a mission. For instance, on 6 June 2025, the Japanese lunar lander Resilience (Hakuto-R Mission 2) had a hard landing during its final descent on the Moon when its laser range finder began outputting erroneous altitude values in the last few kilometers before touchdown~\cite{ispace2025press}. This highlights the need for accurate monitoring systems that can address the problem of fault detection and isolation (FDI) in perception systems.

    \begin{figure}[t!]
        \centering
        \includegraphics[width=0.48\textwidth, clip,keepaspectratio]{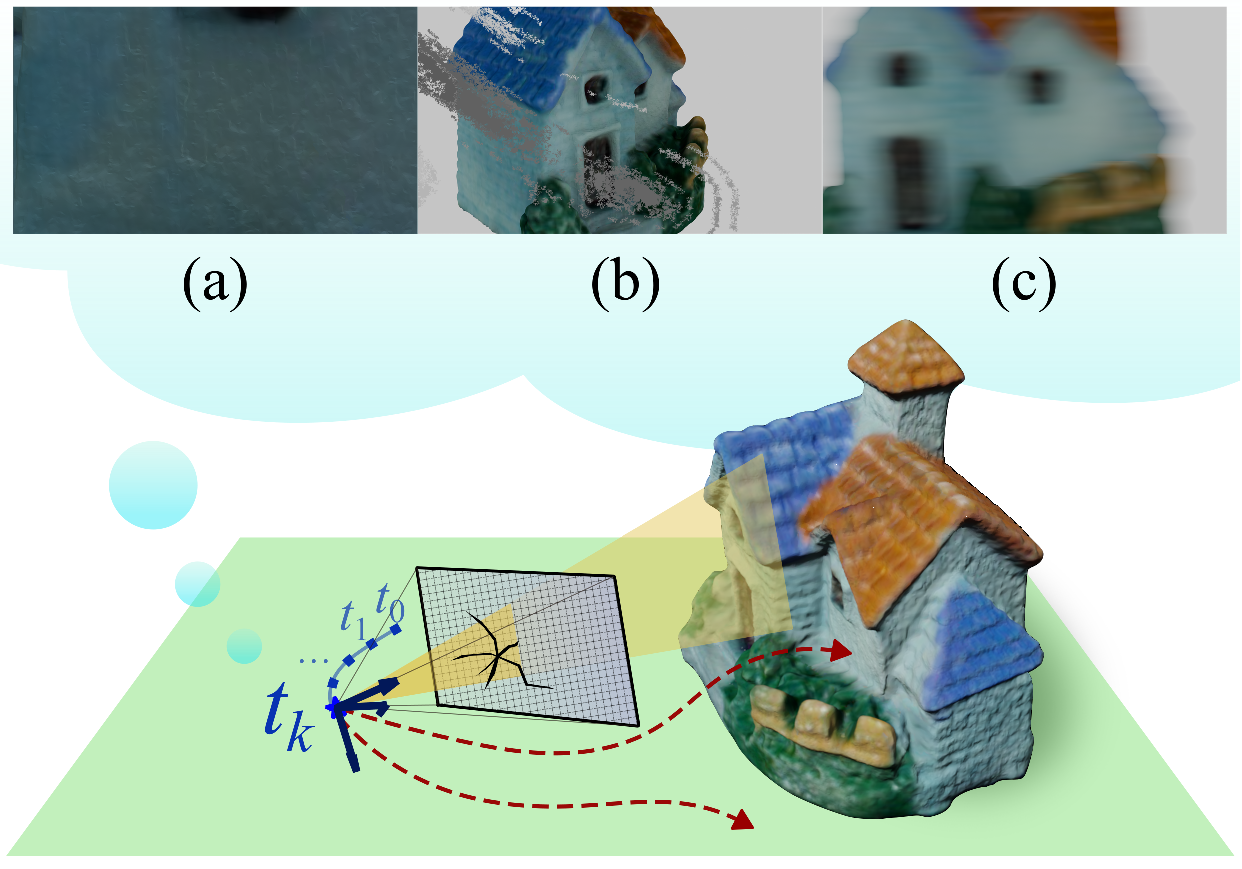}
        \caption{Our FDI method predicts perception outputs under fault hypotheses and exploits control inputs to enhance fault detectability and isolability in robot perception systems. Examples of vision faults are (a) visual deprivation, (b) sensor damage, (c) motion blur. Adapted from OmniObject3D~\cite{wu2023omniobject3d}, licensed under CC BY 4.0 (https://creativecommons.org/licenses/by/4.0/); changes: rendered with Blender.}\label{fig:1}
        \vspace{-4mm}
    \end{figure}

	The method presented in this paper can handle a broad range of fault and failure types, including both physical malfunctions and algorithmic errors in perception systems that cause deviations from their intended functionality. On the physical side, sensors can suffer calibration shifts, temporary occlusions, and environmental interference~\cite{goelles_fault_2020}. At the algorithmic level, deep neural networks (DNN) can misclassify objects due to distribution shifts (i.e., out-of-distribution inputs), and multi-sensor fusion can become erroneous due to calibration issues~\cite{hou_fault_2023, mitra_formal_2025, ji_multi-modal_2021}. Moreover, faults that occur at an early stage of a perception system's pipeline propagate through it and do not remain isolated~\cite{hsieh_assuring_2024}. Hence, FDI methodologies that rely on physical redundancy may not be sufficient~\cite{oconnell_learning-based_2024}. Thus, one needs to study and investigate FDI methodologies based on the available analytical redundancy in perception systems. Once a certain fault is detected and isolated, a fault recovery control can be implemented.

    \begin{figure*}[thpb]
        \centering
        \includegraphics[width=0.9\textwidth, clip,keepaspectratio]{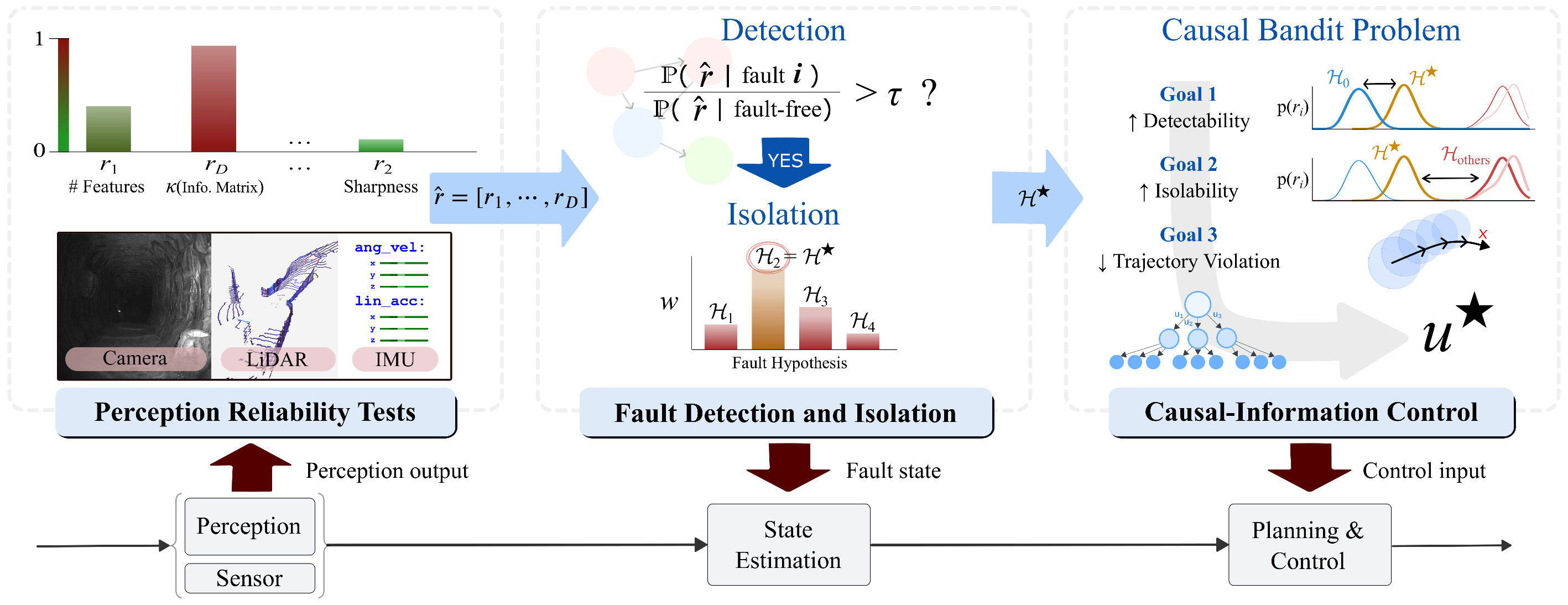}
        \caption{Algorithm architecture overview. The FDI module compares the perception reliability test results with the fault-hypothesis distributions and generates control inputs that improve FDI performance. Camera and LiDAR data captured using Foxglove from CERBERUS DARPA Subterranean Challenge datasets~\cite{https://doi.org/10.48550/arxiv.2207.04914,doi:10.1126/scirobotics.abp9742}}\label{fig:algorithmArchitecture}
        \vspace{-4mm}
    \end{figure*}
    
	\subsection{Related Work}
    The faults that occur in Simultaneous Localization and Mapping (SLAM) and Visual Inertial Odometry (VIO) systems are sensor faults~\cite{antonante_monitoring_2023, goelles_fault_2020}, tracking failures~\cite{rahman_large-scale_2024, hussain_qureshi_one_2024}, data association failures~\cite{pathak_unified_2018, shienman_d2a-bsp_2022}, and filtering inconsistency problems~\cite{zhang_asro-dio_2023}. Sensor faults are caused by hardware damage or software malfunction. Faults in front-end modules, such as tracking and data association failures, are often caused by visually deprived conditions (i.e., textureless surfaces and repetitive patterns), dynamic scenes (i.e., aggressive camera motion), and undesirable lighting conditions (i.e., high-contrast images). Lastly, the filtering inconsistency problems, a type of fault in back-end modules, result from large inter-frame transformations that trigger the accumulation of linearization errors.

    The work in \cite{hou_fault_2023} compares perception outputs with a predefined fault threshold for runtime monitoring. Additionally, \cite{antonante_monitoring_2023} developed fault diagnostic graphs to associate errors with individual perception module outputs, as evaluated by diagnostic tests. Although these works enable FDI, they rely on having redundant sensors, which can be costly. To enhance the robustness of SLAM \cite{zhao_how_2024} developed image quality metrics to select confident features or scenes. Similarly, feature quality metrics that assess keypoint co-visibility between frames~\cite{rahman_large-scale_2024, samadzadeh_srvio_2023, antonante_monitoring_2023} and the dynamic scene metrics that leverage vehicle velocity~\cite{rahman_large-scale_2024}, optical flow~\cite{han_dynamic_2020}, and image sharpness~\cite{guo_deblurslam_2021, min_coeb-slam_2023} have been proposed. 

    
	\subsection{Contributions}
    
    We define perception reliability tests for various fault modes to capture differences between fault-free and fault-induced behaviors. We utilize the structural causal model (SCM) formalism of Pearl \cite{pearl2009causality} and its operational rules for interventions and counterfactual queries, where we treat each hypothesized fault mode as an intervention on the perception pipeline. 
	We then introduce and define an information-theoretic metric based on the Kullback–Leibler (KL) divergence between the reliability test results and those from a baseline fault-free case to measure the detectability and isolability of the hypothesized faults. This metric, designated as Effective Information (EI), captures how control inputs influence the reliability test results by affecting the autonomous system's state. To the best of our knowledge, this is the first work that studies the FDI as a counterfactual reasoning problem for a closed-loop autonomous system and also connects the informativeness of control inputs to the detection and isolation of the hypothesized faults. Finally, we show that finding the control input that helps maximizing the EI leads to having a causal bandit problem~\cite{bareinboim2015bandits}, where each action arm corresponds to an intervention on the control input that improves our FDI accuracy. A Monte-Carlo Tree Search (MCTS) approach with Upper Confidence Bound (UCB)~\cite{kocsis2006bandit,silver2010monte} that penalizes large deviations from primary mission objectives (e.g., tracking a trajectory) is employed to solve the mentioned causal bandit problem.
    
   The main contributions of this paper are as follows.
    \begin{enumerate}
    
        \item We exploit analytical redundancy of the perception system and actively use control inputs for FDI by applying the do-operator from causal inference. This is achieved via a counterfactual reasoning approach, where it is analyzed how control inputs affect reliability test outcomes under various fault hypotheses. A quantitative detection and isolation metric measuring the informativeness of each control input for FDI is introduced. 
    
        \item We formulate the problem of selecting control inputs for FDI as a causal bandit problem. Using a MCTS strategy with UCB, we maximize a weighted reward function that prioritizes inputs informative about the most likely fault modes. In addition, our reward function penalizes large deviations from the desired trajectory of the system.

        \item Our FDI method uses the distribution of reliability test results under various fault modes and accounts for the uncertainty inherent in the perception system's outputs. Thus, our method encodes more information than mean value and threshold-based FDI methods, which only reflect the central tendency of a distribution. 
    
    \end{enumerate}

    \vspace{-0.5mm}
    \section{Preliminaries and Problem Statement}\label{s:prelim}
	\vspace{-0.5mm}
	\subsection{Structural Causal Models and $\mathrm{do}$-operator}
	An SCM~\cite{pearl2009causality} describes a model $M$ using a set of endogenous variables $V=\{X_1,\dots,X_{o}\}$ generated by structural equations $X_i \coloneqq m_i(\mathrm{Pa}_{X_i},\hat{A}_i)$, where $\mathrm{Pa}_{X_i}\subseteq V \setminus \{X_i\}$ are the parents of $X_i$ and $\hat{A}_i$ are exogenous variables, for $i=1,\dots,o$. 
    \begin{definition}[$\mathrm{do}$-operator]\label{def:do_operator}
        An intervention that forcefully sets a subset of variables $X_S\subseteq V$ to values $x_S$ is denoted by the $\mathrm{do}$-operator, i.e., $\mathrm{do}(X_S=x_S)$.
    \end{definition}
    Executing $\mathrm{do}(X_S=x_S)$ removes the structural equations for $X_S$ and replaces them with $X_S=x_S$, which yields the model $M_\text{S}$. Consequently, the post-intervention distribution can be written as $p_{M}(v | \mathrm{do}(x_S))= p_{M_\text{S}}(v)$. In other words, under model $M$, the distribution of outcome $V$ after the intervention $\mathrm{do}(x_S)$ is the probability that model $M_\text{S}$ assigns to $V$. It is worth noting that $p_{M}(v | \mathrm{do}(x_S))$ is different from having $p_{M}(v | x_S)$ in the sense that $\mathrm{do}(x_S)$ breaks incoming causal links into $X_S$ rather than observing their values. 
    
	\subsection{Effect Distribution (ED)}
	Let $\hat{V}$ be a variable of interest and $\hat{A}\in\hat{\mathcal{A}}$ an intervention variable (e.g., the control input, hypothesized faults) with the distribution $p_{\hat{A}}(\hat{a})$. The Effect Distribution (ED)~\cite{chvykov2020causal} is the marginal distribution of $\hat{V}$ induced by randomizing $\hat{A}$ such that
	\begin{align}\label{e:ED_general}
		\text{ED}_{\hat{V}}(\hat{v})= \mathbb{E}_{\hat{a}\sim p_{\hat{A}}}[p(\hat{v} | \mathrm{do}(\hat{a}))]=\int_{\hat{\mathcal{A}}} p(\hat{v} | \mathrm{do}(\hat{a}))p_{\hat{A}}(\hat{a})d\hat{a}.
	\end{align}
	The $\text{ED}_{\hat{V}}$ captures the baseline behavior of	$\hat{V}$ under all interventions we are able to perform on $\hat{A}$.
	
	\subsection{Effective Information (EI)}
	Suppose we actively sample $\hat{U}\sim p_{\hat{U}}$ and record the resulting $\hat{V}$. The Effective Information (EI)~\cite{chvykov2020causal,hoel2017map,tononi2003measuring} from $\hat{A}$ to $\hat{V}$ can be defined as
	\begin{equation}\label{e:EI_general}
		\mathrm{EI}(\hat{A}\rightarrow \hat{V})= \mathbb{E}_{\hat{a}\sim p_{\hat{A}}}[ D_{\mathrm{KL}}\big(	p(\hat{v} | \mathrm{do}(\hat{a})) \Vert\text{ED}_{\hat{V}}(\hat{v})\big)],
	\end{equation}
	where $D_{\mathrm{KL}}$ is the Kullback–Leibler (KL) divergence. Consequently, the EI of an individual $\hat{a}_i\in \hat{A}$ is $D_{\mathrm{KL}}\big(p(\hat{v} | \mathrm{do}(\hat{a}_i)) \Vert\text{ED}_{\hat{V}}(\hat{v})\big)$ which indicates the impact that $\hat{a}_i$ makes in the variable $\hat{v}$. The EI \eqref{e:EI_general} is the average statistical distance (in the sense of the KL divergence) between the	post-intervention distribution and the baseline $\mathrm{ED}_{\hat{V}}$.

    \subsection{System Model}
	We consider the following nonlinear control-affine system:\begin{align}\label{e:sys}
		{x}(t+1) &= f(x(t)) + g(x(t))u(t) + w(t),
	\end{align}
	where $x(t) \in \mathbb{R}^n$ is the state, $u(t) \in \mathbb{R}^m$ is the control input, and $w(t)\in \mathbb{R}^n$ denotes the process noise. Moreover, $f(x)$ and $g(x)$ are smooth functions.

	The perception system integrates $S\in \mathbb{N}^+$ sensors (e.g., camera, LiDAR) in its pipeline, which consists of various layers (e.g., preprocessing, feature extraction), and $\mathbb{N}^+$ is the set of positive integers. The output of the perception system is given by
	\begin{align}\label{e:perception}
		y(t) = (h\circ z)(x(t),e(t))+\nu(t),
	\end{align}
	where $y(t)\in\mathbb{R}^{p}$ is the state to be estimated via the perception system, $z: \mathbb R^n \times \mathbb R^{n_e} \rightarrow\mathbb R^{n_z}$ is the function describing the output of perception sensors, $h: \mathbb R ^{n_z} \rightarrow \mathbb R^p$ is a perception map from the sensor output to $y$, $e(t)\in \mathbb{R}^{n_\text{e}}$ represents unmeasured environmental states (e.g., lighting condition, weather changes, dynamic objects), $\nu(t) \in \mathbb{R}^{p}$ is the measurement noise. 
    The definition of fault mode in perception systems considered in this paper is as follows.

    \begin{definition}[Fault Mode]\label{def:faultMode}
    A fault mode is the function $t\rightarrow \delta_i(t)$ that represents the occurrence of a fault scenario $\mathcal{H}_i$, such that $\delta_i (t) = \mathbf{1}_{\mathcal{H}_i}(t), \quad i=1,\dots,M$, where $\mathbf{1}_{\mathcal{H}_i}(t)\in\{0,1\}$ is the indicator function of the $i$-th fault hypothesis, i.e., $\mathcal{H}_i$, at time $t$, and $M\in\mathbb{N}^+$ is the number of considered fault scenarios.
    \end{definition}

	
	Fault modes are inherently case-specific; representative examples include sensor damage, visually deprived conditions, and dynamic scenes. See Section~\ref{ss:applic} for their relevance in vision-based asteroid exploration of a space robot. We consider the following assumption throughout this paper.

    \begin{assumption}\label{assum:oneFault}
		At any instance of time, there exists only one fault in the perception system \eqref{e:perception}, i.e., the cardinality of $\mathrm{supp}(\delta)=\{i\in\{1,\dots,M\}\, |\, \delta_i\neq0\}$ is equal to $1$.
    \end{assumption}

    For brevity, we omit the explicit dependence of the variables on $t$ in the remainder of the paper.
	
    \subsection{Problem Statement}
    Given the perception system \eqref{e:perception}, we develop and design perception reliability tests based on the system state $x(t)$ and internal variables of the perception system from $(h\circ z)(x(t),e(t))$. In order to avoid the direct comparison of high-dimensional visual information, the outputs of reliability tests are generated for various hypothesized fault scenarios and efficiently compared with observed test results. Consequently, we define a set of counterfactual hypotheses regarding faults in the perception system, i.e., $\delta_i$, to study the detection and isolation of the hypothesized faults. To carry out the latter, we introduce a causal information-theoretic metric, designated as EI, that captures how distinguishable various hypothesized faults are for a certain control input. Finally, we find interventions, i.e., control input $u(t)$, that actively enhance the detection and isolation capabilities of our methodology for the hypothesized faults, and we formulate this problem within the causal bandit framework. Our proposed FDI methodology is shown in Fig.~\ref{fig:algorithmArchitecture}.

    \section{Perception reliability tests and Counterfactual Hypotheses}\label{s:counterFactual}

    In this section, perception reliability tests are introduced as a method to derive essential diagnostics from complex perception outputs. As opposed to a direct comparison of the measured perception output with the nominal fault-free approximation of it, a set of perception reliability tests is developed as a measure of the perception system's well-being. Furthermore, the richness of visual information generated by perception sensors enables one to extract intermediate outputs that are more compact than raw sensor data but more informative than the final output alone. These intermediate outputs can be utilized in the set of reliability tests to carry out FDI for perception systems.

	\subsection{Perception Reliability Tests}
    The perception system first captures visual information from its sensors and subsequently processes this visual information through one or more algorithm modules. Let the output of the $j$-th module of the perception pipeline be $y^{(j)} = h^{(j)} \big(z (x, e)\big) = \mathcal{P}^{(j)}(x, e)\in \mathbb R^{n_{h^{(j)}}} $, where $h^{(j)} : \mathbb R^{n_z} \rightarrow\mathbb R^{n_{h^{(j)}}}$ is the function which maps the sensor output to the intermediate output $y^{(j)}$. Based on this intermediate output, a set of perception reliability tests is defined to evaluate the reliability of the perception output for each test.
    
    \begin{definition}[Perception Reliability Test]\label{def:dtest}
    	A perception reliability test associated with module $j$ in a perception system is a function
        \begin{align*}
            d: \underbrace{\mathbb{R}^{n_{h^{(j)}}} \times \mathbb{R}^{n_{h^{(j)}}} \times \cdots \times \mathbb{R}^{n_{h^{(j)}}}}_{W}\rightarrow\mathbb R,
        \end{align*} which maps output sequence $\{y^{(j)}(t)\}_{t=k+1-W}^{k}$ to a scalar value representing reliability measure.
    \end{definition}
    

    The perception reliability test result $r(x_{k+1-W:k},e) = d \big( \{\mathcal{P}^{(j)} (x_{t},e)\}_{t=k+1-W}^{k} \big) $ quantifies the reliability of the output from a specific sensor or algorithmic module and is characterized by the length $W$-sized output derived from the state and the environmental states.

    The choice of reliability tests is application-specific, but the way the perception reliability test is used for fault diagnosis is general enough to accommodate various types of reliability tests. The following example illustrates the perception reliability tests in practice.

    \begin{example}\label{ex:campipeline}
    Visually deprived conditions can cause tracking or data association failures in vision-based SLAM and VIO systems. A perception reliability test that can verify such conditions involves counting the number of detected feature points in an image (e.g., low when the camera is directed toward a textureless surface). For instance, assuming prior knowledge of keypoint locations, the number of detected feature points $ d(\mathcal{Q})$ can be computed from the state vector and the known keypoints as follows:
        \begin{align}
        	\mathcal{Q}(x, e) &=\{i\in\mathbb{N} : C(e)\Pi (x)q_i\in\mathcal{F}, \;q_i\in \mathcal{M} \subset \mathbb{R}^3\}, \nonumber \\
            d(\mathcal{Q}) &= |\mathcal Q (x, e )|, \label{detectfeature}
        \end{align}
    where $\mathcal Q$ is the set of points' indices inside the field of view $\mathcal F\in \mathbb R^2$ defined with respect to the image plane, $C(e)$ is the feature point selection matrix that depends on the environmental conditions, $\Pi(x)$ is the projection matrix derived from the relative pose, $\mathcal{M}$ is the set of known keypoints, and $|\cdot|$ is the cardinality of the set. 
    \end{example}

 


     As such, perception reliability tests are widely employed in practice, often combined with simple threshold-based scores or statistical evaluations, to verify the quality of visual information and validate the correctness of algorithmic results. Yet, the fundamental distinction is that our work utilizes reliability tests to identify the source of the fault, whereas other works use them to reject anomalous measurements or trigger fallback strategies to maintain safe behavior, regardless of the fault origin. For this purpose, we present a model that explicitly accounts for the fault mode dependency in modular and customizable reliability tests. Such a model bridges practical perception pipelines, including neural network–based approaches, with the proposed information-theoretic framework for fault diagnosis. A detailed example of reliability tests that are used in a camera-based perception pipeline is provided in Section~\ref{s:simu}.


    \begin{figure}[t]
    \centering
        \includegraphics[width=0.4\textwidth, clip,keepaspectratio]{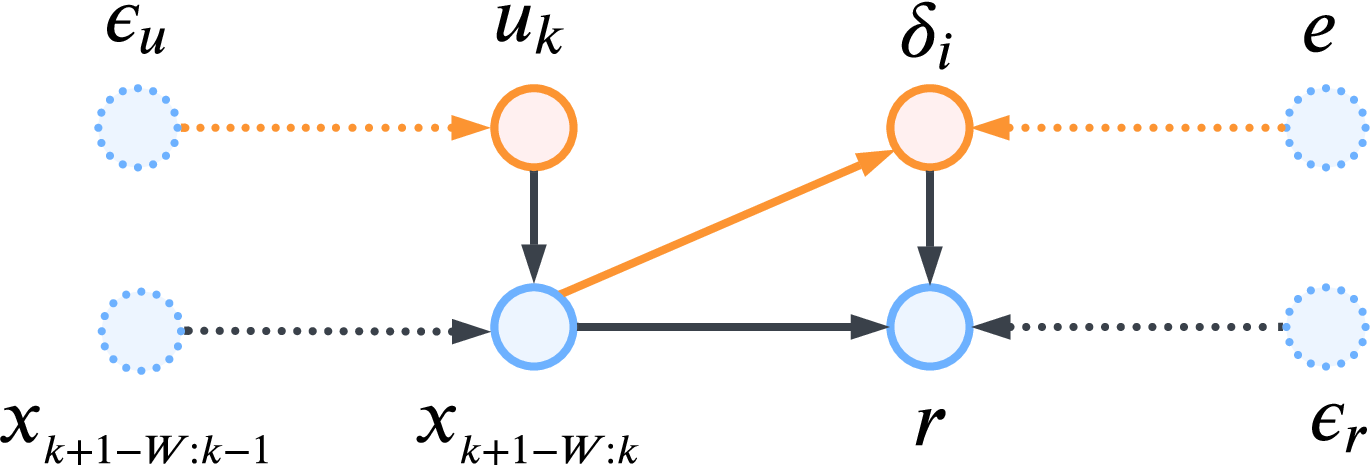}
        \caption{Causal model for perception reliability test with correlation between fault mode, state, and environmental state. The dotted lines indicate the effect of exogenous variables, and the orange lines indicate the links removed after intervention $\mathrm{do}(\delta_j)$ and $\mathrm{do}(u_k)$.}\label{fig:scm}
        \vspace{-4mm}
   \end{figure}
   
	\subsection{Counterfactual Fault Hypotheses} \label{sec:faultH}
	Let us define $r=[r_1, \dots,r_D]^\top\in~\mathbb{R}^D$, where $r_i=d_i \big(\{\mathcal{P}^{(j)} (x_t, e)\}_{t=k+1}^{k+W}\big)$. As described in the previous subsection, each reliability test result $r_i$ is sensitive to a set of fault modes given the state sequence, for $i=1,\dots,D$, where $D$ is the number of tests. This implies that two distinct causes of the fault in the perception system may trigger the same reliability test, potentially leading to fault misclassification.	In order to accurately isolate the faults, we employ the do-operator defined in Section~\ref{s:prelim}, which enables us to treat various hypothetical fault scenarios separately. We consider a healthy scenario and $M$ fault hypotheses in the set $\mathcal{H}=\{\delta_0, \delta_1,\dots,\delta_M\}$, where each $\delta_i$ corresponds to a fault scenario in our perception system, for $i=1,\dots,M$, and $\delta_0$ is the healthy case, i.e., fault-free. 
	
	

    To evaluate the effect of fault modes and control inputs on the reliability test results, a structural causal model $M=\langle E,V,F \rangle$ is defined for the autonomous system at time $t=k$, where $E$ and $V$ denote the sets of exogenous variables and endogenous variables, respectively. Realizations of the endogenous variables are denoted by $u, x_{k+1-W:k},\delta_j, r$, and those of the exogenous variables by $\epsilon_u, \epsilon_{r},e$ with the observed states $x_{k+1-W:k-1}$. The set of functions that maps to $V_i$ from $E\cup V\backslash {V_i}$ are $F=\{F_u, F_{x_{k+1-W:k}}, F_\delta, F_r\}$, with related variables shown in Fig.~\ref{fig:scm}. As per the SCM in Fig.~\ref{fig:scm}, there is $e\to \delta_j\to r$ and no direct $e\to r$ edge. Hence, under the intervention $\mathrm{do}(\delta_j)$, we remove all incoming arrows into $\delta_j$ (e.g., from $x$ and $e$), and any association between $\delta_j$ and the environmental variable $e$ is eliminated. The same method is used for the intervention $\mathrm{do}(u)$.
    
    The effect of intervention due to the fault mode and the control input can be derived using the adjustment~\cite{pearl2009causality}. In our case, the conditional counterfactual due to the intervention $(\mathrm{do}(u),\,\mathrm{do}(\delta_j))$ on the reliability test result $r$ is given by
    \begin{align}
        p_\mathrm{cf}\big (r\lvert &x_{k+1-W:k-1}, \mathrm{do}(u),\mathrm{do}(\delta_j)\big)= \int_{\mathcal{X}}  p(r|x_{k+1-W:k}, \delta_j) \cdot  \nonumber \\ 
        &p(x_{k+1-W:k} \vert x_{k+1-W:k-1}, u) dx_{k+1-W:k}, \label{e:intervention}
    \end{align}
    where $p(r|x_{k+1-W:k}, \delta_j)$ is the reliability test result's distribution given the system state and the fault mode, and $p(x_{k+1-W:k} \vert x_{k+1-W:k-1}, u_{k})$ is the predicted belief at time $k$. The predicted belief can be obtained from the state estimation module, where our state estimator provides the abduction step in Pearl's abduction, action, and prediction counterfactual framework. On the other hand, $p(r|x_{k+1-W:k}, \delta_j)$, which encodes the probabilistic reliability test result under a given state sequence and fault mode, can be obtained from the experiment in controlled environments where the fault scenario has been set up, as we discuss further in Section~\ref{s:simu}. In what follows, all densities and information-theoretic metrics are counterfactual and implicitly conditioned on $x_{k+1-W:k-1}$. We also define $p(r\lvert \mathrm{do}(u),\mathrm{do}(\delta_j)):=p_\mathrm{cf}(r\lvert x_{k+1-W:k-1}, \mathrm{do}(u),\mathrm{do}(\delta_j))$.

	\section{Information-Theoretic Diagnostic Metrics for Fault Detection and Isolation}\label{s:EI}
	Given the defined fault hypotheses in the previous section, we adopt the information-theoretic metric, Effective Information (EI), (see Section~\ref{s:prelim}) and modify it for our FDI problem. The modified EI measures the amount of information that is captured by our reliability test results due to an intervention $\text{do}(u)$ in terms of the detection and isolation of each fault hypothesis. Hence, maximizing the EI would increase the probability of detecting and isolating the hypothesized faults.

	\subsection{Effect Distribution}
	One needs to first establish a baseline for normal behavior by redefining the Effect Distribution (ED) that was introduced in Section~\ref{s:prelim}. We define the ED as the expected distribution of the perception reliability test result vector $r$ under fault-free conditions $\delta_0$, marginalized over interventions in the space of admissible control inputs $\mathcal{U}$. One has
	\begin{equation}\label{e:ED}
		\text{ED}(r|\mathrm{do}(\delta_0))=\mathbb{E}_{u\sim \mathcal{U}} [p(r|\mathrm{do}(u),\mathrm{do}(\delta_0))].
	\end{equation} 
	
	\begin{remark}\label{rem:ED_approx}
		Since the exact computation of $\text{ED}(r)$ is challenging, one can approximate it. Hence, under the assumption of having a Gaussian density for the reliability test distribution under each control input $u$, one has $p(r|\mathrm{do}(u),\mathrm{do}(\delta_0))\approx \mathcal{N}(r;\,\mu,\Sigma)$, where $\mu$ and $\Sigma$ are the mean value and the covariance. Consequently, the $\text{ED}(r)$ in \eqref{e:ED} can be approximated by drawing $\bar{K}$ samples and using a Gaussian Mixture Model (GMM) given by $\text{ED}(r) \approx\frac{1}{\bar{K}}\sum_{k=1}^{\bar{K}} \mathcal{N}(r;\,\mu_k,\Sigma_k)$~\cite{reynolds2009gaussian}. Alternatively, if the assumption of having Gaussian reliability test distributions is restrictive, one can fit a kernel density estimator directly to the $\bar{K}$ samples~\cite{parzen1962estimation}.
	\end{remark}

	\subsection{Fault Detectability Metric}
	The detectability of a fault $\delta_i$ is a measure of how different its signature in the reliability test $r$ is from the fault-free baseline. In order to measure the difference between probability densities, we utilize the KL divergence. Consequently, we define the detectability of the $i$-th hypothesized fault under the control input $u$ in the following form:
	\begin{align}\label{e:detect}
			D_{\text{detect}}^i(u) = D_\mathrm{KL}[p(r | \mathrm{do}(u), \mathrm{do}(\delta_i)) || \, \text{ED}(r)].
	\end{align}
		
	\subsection{Fault Isolation Metric}
	In addition to the detection of faults, our fault diagnosis methodology can isolate various faults in the system. Correct isolation among faults results in accurately identifying which hypothesized fault in the set $\mathcal{H}$ is the true fault in the perception system. Thus, we define the isolation of the $i$-th fault under the control input $u$ as the sum of the KL divergences between the reliability test distribution corresponding to $\delta_i$ and those of all other fault hypotheses, for $j \in \{1,...,M\}$, as given by
	\begin{align}\label{e:isolate}
		D_{\text{isolate}}^i(u) =\sum_{j \neq i} & D_\mathrm{KL}[p(r \mid \mathrm{do}(u), \mathrm{do}(\delta_i))  \lVert \nonumber\\
		& p(r\mid\mathrm{do}(u), \mathrm{do}(\delta_j))].
	\end{align}

	\begin{remark}\label{rem:whyKL}
		It is worth noting that due to the connection of the KL divergence with the notion of mutual information~\cite{cover1999elements}, $D_{\text{detect}}^i(u)$ quantifies how much a perception reliability test distribution $p(r\mid \mathrm{do}(u),\mathrm{do}(\delta_i))$ differs from the baseline ED with respect to the amount of information contained in $r$. Also, the KL divergence is inherently asymmetric, thus, it takes into account the causal direction of interventions. If these considerations are not critical for a particular application, the KL divergence in \eqref{e:detect} and \eqref{e:isolate} can be replaced by an alternative distance measure, such as the Wasserstein metric.
	\end{remark}
    
	\subsection{Effective Information for Detection and Isolation}
	We combine $D_{\text{detect}}^i(u)$ in \eqref{e:detect} and $D_{\text{isolate}}^i(u)$ given by \eqref{e:isolate} into a single metric that captures the total diagnostic value of a control input. The EI of a control input $u$ to identify hypothesis $\delta_i$ is the sum of its detection and isolation metric, as expressed by
	\begin{align}\label{e:EI}
		\text{EI}(u | \delta_i) = D_{\text{detect}}^i(u) + D_{\text{isolate}}^i(u).
	\end{align}
	The EI given in \eqref{e:EI} quantifies how much an active intervention on $u$ reveals diagnostic information in the sense of \eqref{e:ED} and \eqref{e:EI} about the fault $\delta_i$.

	\section{Causal Bandit, Effective Information (EI), and Active Fault Detection and Isolation}\label{s:bandit}
	Considering that EI measures both the detection and isolation of faults given an intervention $\text{do}(u)$, a causal bandit problem~\cite{lattimore2016causal} is studied to maximize it. In order to investigate a solution for the mentioned causal bandit problem, we utilize a Monte Carlo Tree Search (MCTS) algorithm.

	\subsection{Maximizing EI in a Causal Bandit Problem}\label{ss:maxEI}
	Our main objective in this section is to find a control input $u^*$ that maximizes the EI and helps us to isolate the fault in the system. Hence, we define a weighted EI expressed by\begin{align}\label{e:EI_weighted}
		\text{EI}_{\mathrm{w}}(u) = \sum_{i=1}^{M} w_i \cdot \text{EI}(u | \delta_i),
	\end{align}
	where $w_i \in [0, 1]$ such that $\sum_{i=1}^M w_i = 1$. After applying a control action $u$ and observing a new reliability test $r^\prime=r(t+1)$, these weights are updated via Bayes' rule in the following form:
	\begin{align}\label{e:weight}
		w_i(t+1) = \frac{w_i(t) \cdot p(r^\prime | \mathrm{do}(u), \mathrm{do}(\delta_i))}{\sum_{j=1}^{M} w_{j}(t) \cdot p(r^\prime | \mathrm{do}(u), \mathrm{do}(\delta_{j}))}.
	\end{align}
	
	Maximizing the weighted EI in \eqref{e:EI_weighted} by means of the control input $u$ has two main outcomes. First, it drives the system to increase its information in the sense of $D_{\text{detect}}^i(u)$ and $D_{\text{isolate}}^i(u)$. Second, it accelerates the convergence of the hypothesis weights towards the true fault state. The latter is achieved by choosing a control action $u^*$ that maximizes the information gain for the fault hypothesis with the highest weight. Moreover, the Bayesian weight update rule \eqref{e:weight} ensures that the corresponding weight to the true fault hypothesis increases significantly relative to the others. Therefore, the fault can be identified by monitoring the hypothesis weights. Hence, $\hat{i}$ is the index of the identified fault as given by
	\begin{align}
		\hat{i} = {\arg\max}_{i \in \{1,\dots,M\}} w_i.
	\end{align}
	
	The control action that maximizes \eqref{e:EI_weighted} may not be aligned with our control objectives (e.g., tracking a desired trajectory). Therefore, we define the causal bandit problem in the following form:		\begin{align}\label{e:ustar}
		u^* = {\arg\max}_{u \in \mathcal{U}} ( \sum_{i=1}^{M} w_i \cdot \text{EI}(u | \delta_i) - \lambda \cdot P(u) ),
	\end{align}
	where $\lambda > 0$ is a regularization parameter and $P(u) = \max(0, \|x - x_d\| - \epsilon)$ is a penalty for deviating from the reference trajectory $x_d\in \mathbb{R}^n$ beyond a tolerance $\epsilon >0$.

    \subsection{MCTS for Active Fault Detection and Isolation}
   We employ an MCTS-UCB algorithm~\cite{kocsis2006bandit,silver2010monte} to solve the optimization problem \eqref{e:ustar}. The MCTS is effective in handling and balancing the exploration and exploitation of large search spaces. The general MCTS procedure (tree expansion, UCB-based selection, search, simulate, and rollout) follows the standard formulation in~\cite{silver2010monte}. We have implemented the MCTS given in Algorithm~\ref{alg:AFDI}, where the reward used in simulation is aligned with our objective~\eqref{e:ustar}, i.e., $R(u) = \text{EI}_{\mathrm{w}}(u) - \lambda P(u)$. Moreover, for the set of admissible control inputs $\mathcal{U}$, we employ progressive widening to manage tree growth.
   
    \begin{algorithm}
		\caption{Active Fault Detection and Isolation via MCTS}\label{alg:AFDI}
        \small
		\begin{algorithmic}[1]
			\Function{Search}{$x_0, w_0$}
			\State initialize tree $T$ with root history $h=[]$
			\State $N(h) \gets 0$, $Q(h,\cdot)\gets 0$
			\While{planning budget remains}
			\State $R \gets$ \Call{Simulate}{$x_0, w_0, h, 0$}
			\EndWhile
			\State \Return action $u^*$ from the root child maximizing $Q(h,u)$
			\EndFunction

			\Function{Simulate}{$x, w, h, d$}
			\If{$d = k$}
			\State \Return $0$  
			\EndIf
			
			\If{$h \not\in T$}
			\State add node $h$ to $T$, set $N(h)\gets0$, $Q(h,\cdot)\gets0$
			\State initialize empty child set $\mathrm{Children}(h)=\{\emptyset\}$
			\State \Return \Call{Rollout}{$x,w,h,d$}
			\EndIf
			
			\If{$|\mathrm{Children}(h)| < [N(h)^\alpha]$}
			\State sample new action $u_{\mathrm{new}} \sim \mathcal{U}$
			\State add $u_{\mathrm{new}}$ to $\mathrm{Children}(h)$,
			initialize $N(h,u_{\mathrm{new}})=0$, $Q(h,u_{\mathrm{new}})=0$
			\EndIf
			
			\State choose $u' = {\arg\max}_{u\in\mathrm{Children}(h)} Q(h,u) + c\sqrt{\ln N(h)/N(h,u)}$
			\State $x' \gets $ predict system output for $(x,u')$
			\State $r \gets EI_{w'}(u') - \lambda\,P(u')$
			\State $R \gets r + \gamma$\Call{Simulate}{$x',w',h+[u'],d+1$}
			
			\State $N(h,u')\gets N(h,u')+1$
			\State $Q(h,u')\gets Q(h,u') + (R- Q(h,u'))/N(h,u')$
			\State \Return $R$
			\EndFunction

			\Function{Rollout}{$x, w, h, d$}
			\If{$d = k$}
			\State \Return $0$
			\EndIf
			\State sample $u' \sim \mathcal{U}$
			\State $x' \gets$ predict system output for $(x,u')$
			\State $r \gets EI_{w'}(u') - \lambda\,P(u')$
			\State \Return $r + \gamma\,$\Call{Rollout}{$x',w',h+[u'],d+1$}
			\EndFunction
		\end{algorithmic}
    \end{algorithm}
    \vspace{-6mm}

        \subsection{Passive Fault Detection and Isolation}\label{ss:passive_FDI}
        Considering that in certain applications actively modifying control inputs for FDI may be undesirable, we develop a passive methodology in this section. In the passive case, a fault is detected by evaluating the likelihood of an observed reliability test $r'$ under a certain fault hypothesis. We adopt a likelihood ratio test to compare the fault-free hypothesis $\mathcal{H}_0$ with the fault hypothesis $\mathcal{H}_j$, as follows:
        \begin{align*}
            j^* &={\arg\max}_j \frac{p\big (r' \,\lvert \,\mathrm{do}(\delta_j),\mathrm{do}( u)\big)}{p\big (r' \,\lvert \,\mathrm{do}(\delta_0),\mathrm{do}( u)\big) }.
        \end{align*}
        A fault is then declared when the maximum likelihood ratio exceeds a threshold $ \tau$, which triggers the subsequent fault isolation algorithm based on weight updates as explained in Section~\ref{ss:maxEI}. This maximum likelihood ratio test is applied at every FDI time step to monitor the existence of faults.

    \section{Simulation Experiments}\label{s:simu}
    \subsection{Application to Vision-based Asteroid Exploration}\label{ss:applic}
\begin{figure*}[thpb]
    \centering
        \includegraphics[width=0.92\textwidth, clip,keepaspectratio]{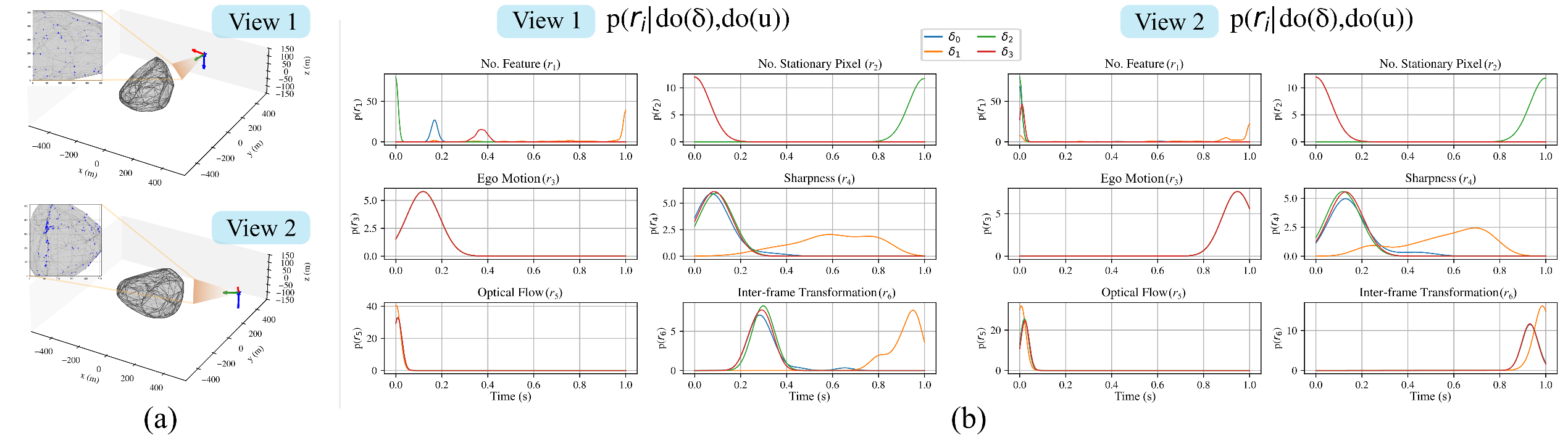}
        \vspace{-2mm}
        \caption{Probability density function of normalized reliability test results for two different relative states. The distribution of $p(r| \text{do}(\delta_j), \text{do}(u))$ varies with scene context and ego motion, even under the same fault hypothesis. (a) Spacecraft's relative position and attitude with respect to the asteroid, (b) probability density of the normalized reliability test results for each case.}
        \label{fig:pdf}
        \vspace{-4mm}
    \end{figure*}

    This section illustrates the application of our fault diagnosis algorithm, where a spacecraft equipped with a monocular camera observes an asteroid and provides relative pose measurements. The spacecraft flies around a Sun-terminator orbit, and the target asteroid is modeled after 99942 Apophis using the publicly available shape model~\cite{noauthor_99942_nodate}. The spacecraft maintains a default nadir-pointing attitude, pointing its camera toward the asteroid's center. Fig.~\ref{fig:pdf} (a) shows the simulation setup used in this study.

    We assume the spacecraft employs vision-based perception for relative navigation during close-proximity operations near the asteroid. The vision-based perception framework used in this simulation consists of image formation, front-end, and back-end modules. The image formation module generates images from the relative states and known keypoints using a geometric vision model with a pinhole camera, the front-end extracts pixel locations of these keypoints, and the back-end estimates the relative pose of the spacecraft from 2D-3D point correspondence, using a Perspective-n-Point (PnP) solver with an Extended Kalman Filter (EKF).

     We evaluate our passive and active FDI methods under multiple perception fault scenarios with this setup.  The fault for this system is defined based on the sensor type (i.e., camera) and the characteristics of the environment (i.e., high-contrast lighting and comparatively slow target dynamics). Accordingly, we define three types of fault: dynamic scene ($\mathcal{H}_1$), sensor damage ($\mathcal{H}_2$), and visually deprived conditions, ($\mathcal{H}_3$), which have corresponding fault modes $\delta_1$, $\delta_2$, and $\delta_3$, respectively. The test scenarios are as follows:
    \begin{itemize}
        \item Case 1: The camera is rotating along the camera principal axis due to disturbance torque.
        \item Case 2: 1.5 seconds after deployment, the camera lens is broken and locally generates random pixel values.
        \item Case 3: The spacecraft passes a region with limited visual information, while the outermost regions of the asteroid along the Sun-terminator orbit normal are unobservable due to extreme lighting conditions.
    \end{itemize}
    
     The fault scenarios to construct the probability distribution are generated from environmental conditions with (i) high relative angular velocity of the scene, (ii) a large number of occluded and wrongly placed pixels, and (iii) undetected pixels in high-illumination and low-illumination regions, for $\mathcal{H}_1$, $\mathcal{H}_2$, and $\mathcal{H}_3$, respectively. To build a probability density function for each hypothesis, we sample 30 fault scenarios for each hypothesis, given the state and the control input.

    \begin{table}[tpb]
    \caption{Reliability tests used in asteroid exploration simulation}
    \label{tab:rel_test_application}
    \begin{center}
    \vspace{-3mm}
    \begin{tabular}{@{}l|l@{}}
    \toprule
    \textbf{Name} & \textbf{Reliability Tests}   \\
    \midrule
    No. Feat. Points 
    & $s_1 = |\mathcal Q (x_k , e )|$     \\
    No. Station. Pix.
    & $s_2 = |\mathcal Q (x_{k-1} , e ) \cap Q (x_k, e )|$  \\
    Ego Motion
    & $s_3 = \lVert v_k \rVert + \sqrt{2} H \lVert  \omega_k \rVert$   \\
    Sharpness
    & $s_4 = \underset{i \in \mathcal{Q}(x_k,e)}{\sum}\frac{H_{\mathrm{blur},i} v_{\mathrm{app},i}}{s_1}$    \\ 
    Optical Flow
    & $s_5 = \underset{{i\in \mathcal{Q}(x_{k-1},e) \cap \mathcal{Q}(x_{k},e)}}{\sum} \frac{\lVert \big(\Pi(x_k) - \Pi(x_{k-1})\big) q_i \rVert }{s_1}$ \\
    Inter-frame TF.
    & $s_6 = \lVert  x_k \ominus x_{k-1} \rVert$ \\
    \bottomrule
    \end{tabular}
    \vspace{-5mm}
    \end{center}
    \end{table}

    The perception reliability tests considered in this case study include: the number of detected features, the number of stationary features, weighted ego motion intensity, image sharpness, average of optical flow, and inter-frame transformation. The reliability tests are defined in the Table~\ref {tab:rel_test_application}, where $x_k=[p_k^\top, v_k^\top, \bar{q}_k^\top, \omega_k^\top]^\top$ is the state vector consisting of relative position, velocity, attitude, angular velocity,  $\mathcal{Q}(x, e)$ is the set of inlier indices as defined in \eqref{detectfeature}, $H$ is the height of the image in terms of pixels, $f$ is the focal length, $t_\mathrm{exp}$ is the exposure time, and $\ominus$ is the state difference operator that handles Euclidean components and quaternion. For calculating the sharpness, we use the following terms:     
    \begin{align*}
        \quad H_{\mathrm{blur},i} &= \frac{f t_\mathrm{exp}}{z_{c,i}^2} \begin{bmatrix} z_c & 0 & -x_c \\ 0 & z_c & -y_c \end{bmatrix}, \\
        \quad v_{\mathrm{app},i} &= T(\bar q_k)  \big(-v_k - (q_i+T(\bar q_k)^\top p_k)\times \omega_k \big),
    \end{align*}
    where $(x_c, y_c, z_c)$ are the feature point location in camera frame and $T(\bar q_k)$ is the direction cosine matrix.

    
    In addition, each reliability test is normalized to produce higher values under faults to ensure stable fault diagnosis. The first test---where higher $s_1$ indicates healthy---is normalized as $r_1 = 1-\tanh({s_1}/{c_1} )$, and the others---where lower values are healthy---are normalized as $r_j =\tanh({s_j}/{c_j}), \; j=2,\cdots, 6$. Here, $c_1,\cdots c_6$ are scaling factors that emphasize the fault-relevant ranges of $s_1,\cdots ,s_6$, hence, improving the sensitivity of FDI.

    The calculation of effect of intervention follows directly from \eqref{e:intervention}. For each FDI step, we compute $p(r|x_{k+1-W:k}, \delta_j)$ with $W=2$ by sampling $x_k\sim p_{X_k}$ and evaluate $r_i$ under each fault hypothesis across 30 randomized scenarios, as if obtained from offline experiments.

    For active FDI, assuming the asteroid's rotation is negligible relative to the spacecraft's motion, we adopt quaternion-based attitude control, with rotational dynamics expressed as $\dot \omega = J_c ^{-1} \big( u_{rot} - \omega \times (J_c \omega)\big)$, where $J_c$ is the spacecraft's moment of inertia, $\omega$ is spacecraft's angular velocity, and $u_{rot}$ is the attitude control input. Specifically, the control input along the camera's $x-$ and $z-$axes is perturbed, while the $y-$axis is constrained only for asteroid pointing. For additional simplicity, the spacecraft's actuation frame is assumed to be aligned with the camera axis.

    \subsection{Simulation Setup}
   The proposed FDI algorithm is implemented in Python, employing PyTorch3D for rendering, kernel density estimation to model reliability distributions, and MCTS for active FDI with rollouts based on prior control inputs. Control inputs are discretized within the reaction wheel torque limits.

    \begin{figure*}[thpb]
    \centering
        \includegraphics[width=0.86\textwidth]{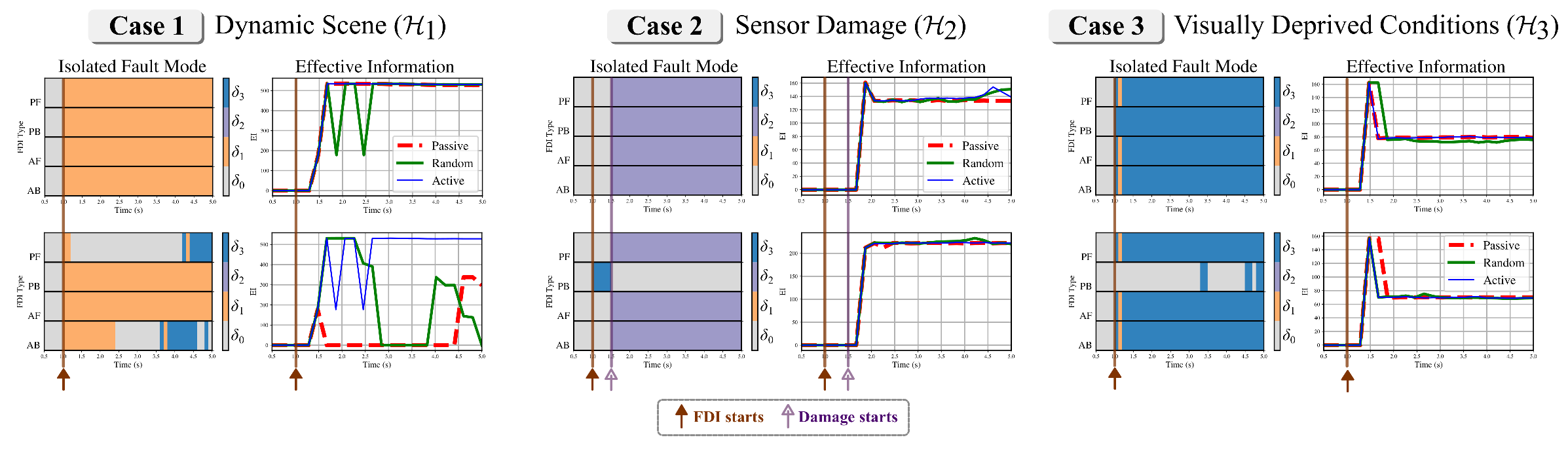}
        \vspace{-4mm}
        \caption{FDI and EI comparison between different methods. The diagnosis process is initiated after 1 second. The upper row represents a severe fault, and the lower row represents a mild fault. PF is our Passive FDI, FB is the Passive Baseline, AF is our Active FDI, AB is the Active Baseline with random control input. The blue curve, corresponding to AF, demonstrates that MCTS-based control input yields higher EI, indicating improved fault diagnosability.}
        \label{fig:fdi_tot}
        \vspace{-5mm}
    \end{figure*}

    \subsection{Reliability Test Distribution}
    The first simulation is designed to verify whether the distribution of the reliability test depends on the state, thereby demonstrating that having a fixed threshold for the reliability test result can lead to false alarms. 
    Figure~\ref{fig:pdf} describes that under fault interventions, the distribution of normalized reliability tests varies with scene context and robot motion. Across two exemplar views, the fault-free hypothesis concentrates most test values near zero, whereas each fault hypothesis typically activates multiple components of ${r}$ with distinct sensitivities, enabling fault isolation: (i) dynamic scenes ($\delta_1$) reduce the number of detected features (increasing $r_1$) and increase motion-related components---$r_3, r_5, r_6$; (ii) sensor damage ($\delta_2$) induces increase in stationary pixels ($r_2$) and broadly degrades reliability; and (iii) visually deprived conditions ($\delta_3$) decrease the number of detected features (increasing $r_1$) while leaving motion-related components relatively healthy. Nevertheless, even under the same fault hypothesis, the induced distributions differ across states, with Views 1 and 2 showing distinct patterns. This motivates active perception FDI: by adjusting the relative state via control, we can elicit more informative patterns in {r} to improve fault detectability and isolability.

    \subsection{FDI Performance Analysis}
    The second experiment is to compare our passive and active FDI methods with baseline FDI methods: threshold-based passive FDI and random control input-based FDI. The threshold-based passive FDI is defined as follows:
    \begin{align*}
        \delta_i=&\delta  _1  \mathbf{1}_{\{r_6>0.3\}}+\delta_2  \mathbf{1}_{\{r_6\le0.3,r_2>0.5\}}+ \\
        &\delta_3 \mathbf{1}_{\{r_6\le0.3,r_2\le0.5,r_1>0.02\}}+\delta_0 \mathbf{1}_{\{\mathrm{otherwise}\}}.
    \end{align*}

    Figure~\ref{fig:fdi_tot} shows the FDI results and EI values for each case, with the upper and lower rows corresponding to severe and mild faults, respectively. Severe faults in this simulation mean faults that induce larger 2-norm estimation errors in the perception system.  In the severe fault cases, all FDI methods detect and isolate the fault within 0.4 seconds of onset, as indicated by the isolated fault mode plots. Passive and active FDI achieve comparable EI values, implying that diagnosability and isolability are already sufficient in these conditions.
    In the mild fault cases, only the active FDI method successfully diagnoses the fault mode while driving the system toward maximum EI. The threshold-based FDI is highly sensitive to scene context, as seen in performance variations across case 3, while our passive FDI fails to detect motion-related faults as in case 1. Although introducing ego-motion can improve performance in such cases, random control inputs cannot maximize EI and result in degraded performance compared to our active FDI method.

    \section{Conclusion}\label{s:conclu}
    This paper presented a perception-monitoring FDI framework based on counterfactual reasoning. The method utilizes intermediate perception outputs and reliability tests to extract diagnostic information without relying on sensor redundancy, and fault hypotheses are evaluated through distributions derived from the fault and control interventions. To improve FDI performance, we introduced EI, which quantifies fault detectability and isolability via the KL divergence. Control inputs are optimized via solving a causal bandit problem using MCTS, to maximize EI while maintaining trajectory tracking.
    Results of simulation for an asteroid exploration scenario show that counterfactual reasoning combined with active control can render perception systems both more robust and self-explanatory. Beyond this, it establishes causal inference as a rigorous formalism to detect and isolate faults in complex perception pipelines. 

\bibliographystyle{ieeetr}
\bibliography{bib/lib_combine} %
        
\end{document}